\documentclass[10pt,twocolumn,letterpaper]{article}

\usepackage{cvpr}
\usepackage{times}
\usepackage{epsfig}
\usepackage{graphicx}
\usepackage{amsmath}
\usepackage{amssymb}
\usepackage[T1]{fontenc}
\usepackage[utf8]{inputenc}
\usepackage{authblk}
\usepackage{lipsum}

\usepackage[breaklinks=true,bookmarks=false]{hyperref}

\cvprfinalcopy 

\def\cvprPaperID{9537} 

\newcommand\blfootnote[1]{%
  \begingroup
  \renewcommand\thefootnote{}\footnote{#1}%
  \addtocounter{footnote}{-1}%
  \endgroup
}
\begin{document}

\title{Towards Practical Lottery Ticket Hypothesis for Adversarial Training}

\makeatletter
\renewcommand\AB@affilsepx{, \protect\Affilfont}
\makeatother

\author[*1]{Bai Li}
\author[*2]{Shiqi Wang}
\author[3]{Yunhan Jia}
\author[4]{Yantao Lu}
\author[5]{Zhenyu Zhong}
\author[1]{Lawrence Carin}
\author[2]{Suman Jana}
\affil[1]{Duke University}
\affil[2]{Columbia University}
\affil[3]{University of Michigan}
\affil[4]{Syracuse University}
\affil[5]{Baidu USA}

\renewcommand\Authands{ and }

\maketitle

\begin{abstract}
\label{sec:abstract}

Recent research has proposed the lottery ticket hypothesis, suggesting that for a deep neural network, there exist trainable sub-networks performing equally or better than the original model with commensurate training steps. While this discovery is insightful, finding proper sub-networks requires iterative training and pruning. The high cost incurred limits the applications of the lottery ticket hypothesis. We show there exists a subset of the aforementioned sub-networks that converge significantly faster during the training process and thus can mitigate the cost issue.
We conduct extensive experiments to show such sub-networks consistently exist across various model structures for a restrictive setting of hyperparameters ($e.g.$, carefully selected learning rate, pruning ratio, and model capacity). 
As a practical application of our findings, we demonstrate that such sub-networks can help in cutting down the total time of adversarial training, a standard approach to improve robustness, by up to 49\% on CIFAR-10 to achieve the state-of-the-art robustness. 
\end{abstract}
\blfootnote{*Equal contribution. This work was done during an internship at Baidu USA.}

\label{introduction}
\section{Introduction}
Pruning has served as an important technique for removing redundant structure in neural networks \cite{han2015learning,han2015deep,li2016pruning,he2017channel}. Properly pruning can reduce cost in computation and storage without harming performance. However, pruning was until recently only used as a post-processing procedure, while pruning at initialization was believed ineffective \cite{han2015deep,li2016pruning}. Recently, \cite{frankle2018the} proposed the lottery ticket hypothesis, showing that for a deep neural network there exist sub-networks, when trained from certain initialization obtained by pruning, performing equally or better than the original model with commensurate convergence rates. Such pairs of sub-networks and initialization are called winning tickets. 

This phenomenon indicates it is possible to perform pruning at initialization. However, finding winning tickets still requires iterative pruning and excessive training. Its high cost limits the application of winning tickets.
Although \cite{frankle2018the} shows that winning tickets converge faster than the corresponding full models, it is only observed on small networks, such as a convolutional neural network (CNN) with only a few convolution layers. In this paper, we show that for a variety of model architectures, there consistently exist such sub-networks that converge significantly faster when trained from certain initialization after pruning. We call these \textit{boosting tickets}.

We observe the standard technique introduced in~\cite{frankle2018the} for identifying winning tickets does not always find boosting tickets. In fact, the requirements are more restrictive. We extensively investigate underlining factors that affect such boosting effect, considering three state-of-the-art large model architectures: VGG-16 \cite{simonyan2014very}, ResNet-18 \cite{he2016deep}, and WideResNet \cite{zagoruyko2016wide}. We conclude that the boosting effect depends principally on three factors: ($i$) learning rate, ($ii$) pruning ratio, and ($iii$) network capacity; we also demonstrate how these factors affect the boosting effect. By controlling these factors, after only one training epoch on CIFAR-10, we are able to obtain 90.88\%/90.28\% validation/test accuracy (regularly requires $>$30 training epochs) on WideResNet-34-10 when 80\% parameters are pruned.

We further show that the boosting tickets have a practical application in accelerating adversarial training, an effective but expensive defensive training method for obtaining robust models against adversarial examples. Adversarial examples are carefully perturbed inputs that are indistinguishable from natural inputs but can easily fool a classifier \cite{szegedy2013intriguing, goodfellow2014explaining}. 

We first show our observations on winning and boosting tickets extend to the adversarial training scheme.
Furthermore, we observe that the boosting tickets pruned from a weakly robust model can be used to accelerate the adversarial training process for obtaining a strongly robust model. On CIFAR-10 trained with WideResNet-34-10, we manage to save up to 49\% of the total training time (including both pruning and training) compared to the regular adversarial training process.
Our code is available at \url{https://github.com/boosting-ticket/ticket_robust}.


Our contributions are summarized as follows:
\begin{enumerate}
\setlength\itemsep{0.3em}
    \item We demonstrate that there exists boosting tickets, a special type of winning tickets that significantly accelerate the training process  while still maintaining high accuracy.
    \item We conduct extensive experiments to investigate the major factors affecting the performance of boosting tickets.
    \item We demonstrate that winning tickets and boosting tickets exist for adversarial training scheme as well. 
    \item We show that pruning a non-robust model allows us to find winning/boosting tickets for a strongly robust model, which enables accelerated adversarial training process.
\end{enumerate}

\label{sec:background}
\section{Background and Related Work}

In this section, we give a brief overview of several topics that are closely related to our work. 
\subsection{Network Pruning}
Network pruning has been extensively studied as a method for compressing neural networks and reducing resource consumption. \cite{han2015learning} propose to prune the weights of neural networks based on their magnitudes. Their pruning method significantly reduces the size of neural networks and has become the standard approach for network pruning. This type of approach is also referred to as unstructured pruning, where the pruning happens at individual weights \cite{han2015deep}. In contrast, structured pruning aims to remove whole convolutional filters or channels \cite{li2016pruning,he2017channel}. While structured pruning often yields better model compression and acceleration without utilizing special hardware or libraries, it can hardly retain the same performance as the full models when the proportion of pruned weights is large.
Besides the magnitude-based pruning strategies, there also exist other types of pruning algorithms like dynamic surgery~\cite{guo2016dynamic}, incorporating sparse constraints~\cite{zhou2016less}, and optimal brain damage~\cite{lecun1990optimal}. However, magnitude-based pruning is more stable on different pruning tasks. Therefore, in this work, we focus on unstructured pruning based on magnitudes of the weights.

\subsection{Lottery Ticket Hypothesis}


Surprisingly, recent research has shown it is possible to prune a neural network at the initialization and still reach similar performance as the full model \cite{liu2018rethinking,lee2018snip}. Within this category, the lottery ticket hypothesis \cite{frankle2018the} states a randomly-initialized dense neural network contains a sub-network that is initialized such that, when trained in isolation, learns as fast as the original network and matches its test accuracy.
    
In \cite{frankle2018the}, an iterative pruning method is proposed to find such sub-networks. Specifically, this approach first randomly initializes the model. The initialization is stored separately and the model is trained in the standard manner until convergence. Then a certain proportion of the weights with the smallest magnitudes are pruned while remaining weights are reset to the previously stored initialization and ready to be trained again. This train-prune-reset procedure is performed several times until the target pruning ratio is reached. Using this pruning method, they show the resulting pruned networks can be trained to similar accuracy as the original full networks, which is better than the model with the same pruned structure but randomly initialized.

Although the lottery ticket hypothesis has been extensively investigated in the standard training setting \cite{frankle2018the,frankle2019lottery}, little work has been done in the adversarial training scheme. A recent work \cite{ye2019second} even argues lottery ticket hypothesis fails to hold when adversarial training is used. In this paper, we show lottery ticket hypothesis still holds for adversarial training and explain the reason why \cite{ye2019second} failed.
 
One of the limitations of the lottery ticket hypothesis, as pointed in \cite{frankle2019lottery}, is that winning tickets are found by unstructured pruning which does not necessarily yield faster training or executing time. In addition, finding winning tickets requires training the full model beforehand, which is time-consuming as well, especially considering iterative pruning. In this paper, we are managed to find winning tickets with much lower time consumption while maintaining the superior performance.

\subsection{Adversarial Examples}

Given a classifier $f:\mathcal{X}\rightarrow \{1,\dots,k\}$ for an input $\mathbf{x}\in \mathcal{X}$, an adversarial example $\mathbf{x}_{\text{adv}}$ is a perturbed version of $\mathbf{x}$ such that $\mathcal{D}(\mathbf{x},\mathbf{x}_{\text{adv}})<\epsilon$ for some small $\epsilon>0$, yet being mis-classified as $f(\mathbf{x})\neq f(\mathbf{x}_{\text{adv}})$. $\mathcal{D}(\cdot, \cdot)$ is some distance metric which is often an $\ell_p$ metric, and in most of the literature $\ell_\infty$ metric is considered, so as in this paper. 

The procedure of constructing such adversarial examples is often referred to as adversarial attacks. 
One of the simplest attacks is a single-step method, Fast Gradient Sign Method (FGSM) \cite{goodfellow2014explaining}, manipulating inputs along the direction of the gradient with respect to the outputs: $$\mathbf{x}_{\text{adv}}=\Pi_{\mathbf{x}+\mathcal{S}}(\mathbf{x}+\alpha (\nabla_\mathbf{x}L(\theta,\mathbf{x},y))$$

where $\Pi_{\mathbf{x}+\mathcal{S}}$ is the projection operation that ensures adversarial examples stay in the $\ell_p$ ball $\mathcal{S}$ around $\mathbf{x}$. Although this method is fast, the attack is weak and can be defended easily. On the other hand, its multi-step variant, Projected Gradient Descend (PGD), is one of the strongest attacks \cite{kurakin2016adversarial,madry2017towards}: $$\mathbf{x}^{t+1}_{\text{adv}}=\Pi_{\mathbf{x}+\mathcal{S}}(\mathbf{x}^t_{\text{adv}}+\alpha (\nabla_\mathbf{x}L(\theta,\mathbf{x},y))$$
where $\mathbf{x}$ is initialized with a random perturbation. Since PGD requires to access the gradients for multiple steps, it will incur high computational cost.


On the defense side, currently the most successful defense approach is constructing adversarial examples via PGD during training and add them to the training sets as data augmentation, which is referred to as adversarial training \cite{madry2017towards}. 

The motivation behind is that finding a robust model against adversarial examples is equivalent to solving the saddle-point problem  $\min_\theta\max_{\mathbf{x}^\prime: D(\mathbf{x},\mathbf{x}^\prime)<\epsilon} L(\theta,\mathbf{x}^\prime,y)$. The inner maximization is equivalent to constructing adversarial examples, while the outer minimization performs as a standard training procedure for loss minimization. 

One caveat of adversarial training is its computational cost due to performing PGD attacks at each training step. Alternatively, using FGSM during training is much faster but the resulting model is robust against FGSM attacks but vulnerable against PGD attacks~\cite{kurakin2016adversarial}. In this paper, we show it is possible to combine the advantages of both and quickly train a strongly robust model benefited from the boosting tickets. 

\subsection{Connecting robustness and compactness}

Prior studies have shown success in achieving both compactness and robustness of the trained networks~\cite{guo2018sparse,ye2019second,zhao2018compress,dhillon2018stochastic,sehwag2019towards,wijayanto2019towards}. However, most of them will either incur much higher training cost or sacrifice robustness from the full model. On the contrary, our framework only requires comparable or even reduced training time than standard adversarial training while obtaining similar/higher robust accuracy than the original full network.


\section{Empirical Study of Boosting Tickets}
\label{boost}
\noindent\textbf{Setup.} As we introduce our methods or findings through experimental results, we first summarize the setup for our experiments. We use BatchNorm \cite{batchnorm}, weight decay, decreasing learning rate schedules ($\times 0.1$ at $50\%$ and $75\%$), and augmented training data for training models. We try to keep the setting the same as the one used in \cite{frankle2018the} except we use one-shot pruning instead of iterative pruning. It allows the whole pruning and training process to be more practical in real applications. On CIFAR-10 dataset, we randomly select 5,000 images out of 50,000 training set as validation set and train the models with the rest. The reported test accuracy is measured with the whole testing set. 

All of our experiments are run on four Tesla V100s, 10 Tesla P100s, and 10 2080 Tis. For all the time-sensitive experiments like adversarial training on WideResNet-34-10 in Section~\ref{subsec:wrn}, we train each model on two Tesla V100s with data parallelism. For the rest ones measuring the final test accuracy, we use one gpu for each model without parallelism. 

We first investigate boosting tickets on the standard setting without considering adversarial robustness. In this section, we show that with properly chosen hyperparameters, we are managed to find boosting tickets on VGG-16 and ResNet that can be trained much faster than the original dense network. Detailed model architectures and the setup can be found in Supplementary Section A.

\subsection{Existence of Boosting Tickets}
To find the boosting tickets, we use a similar algorithm for finding winning tickets, which is briefly described in the previous section and will be detailed here. First, a neural network is randomly initialized and saved in advance. Then the network is trained until convergence, and a given proportion of weights with the smallest magnitudes are pruned, resulting in a mask where the pruned weights indicate 0 and remained weights indicate 1. Unless specified, we always prune the smallest $80\%$ weights, that is the pruning ratio is $80\%$. We call this train-and-prune step \textit{pruning}. This mask is then applied to the saved initialization to obtain a sub-network, which are the boosting tickets. All of the weights that are pruned (where zeros in the mask) will remain to be 0 during the whole training process. Finally, we can retrain the sub-networks.

The key differences between our algorithm and the one proposed in \cite{frankle2018the} to find winning tickets are ($i$) we use a small learning rate for pruning and retrain the sub-network (tickets) with learning rate warm-up from this small learning rate. In particular, for VGG-16 we choose 0.01 for pruning and warmup from 0.01 to 0.1 for retraining; for ResNet-18 we choose 0.05 for pruning and warmup from 0.05 to 0.1 for retraining; ($ii$) we find it is sufficient to prune and retrain the model only once instead of iterative pruning for multiple times. In Supplementary Section B, we show the difference of boosting effects brought from the tickets found by iterative pruning and one-shot pruning is negligible. Note warmup is also used in \cite{frankle2018the}. However, they propose to use warmup from small learning rate to a large one during pruning as well, which hinders the boosting effect as shown in the following experiments.

\begin{figure}[ht]
    \centering
    \includegraphics[width=0.45\textwidth]{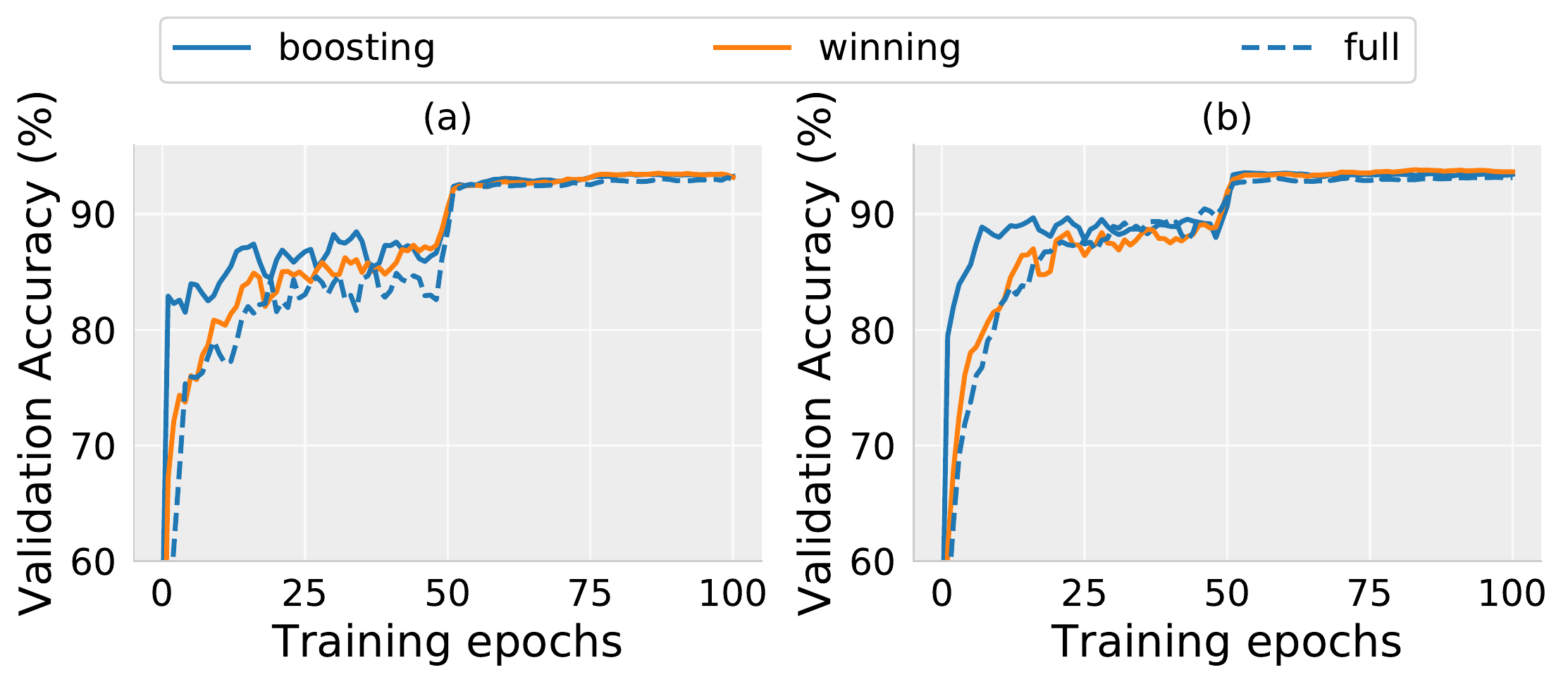}
    \caption{Validation accuracy during the training process on VGG-16 (a) and ResNet-18 (b) for winning tickets, boosting tickets, and the original full models. In both models, the boosting tickets show faster convergence rate and equally good performance as the winning tickets.}
    \label{fig:vgg-resnet}
\end{figure}

First, we show the existence of boosting tickets for VGG-16 and ResNet-18 on CIFAR-10 in Figure \ref{fig:vgg-resnet} and compare to the winning tickets. In particular, we show the boosting tickets are winning tickets, in the sense that they reach comparable accuracy with the original full models. When compared to the winning tickets, boosting tickets demonstrate equally good performance with a higher convergence rate. Similar results on MNIST can be found in Supplementary Section C.

To measure the overall convergence rate, early stopping seems to be a good fit in the literature. It is commonly used to prevent overfitting and the final number of steps are used to measure convergence rates. However, early stopping is not compatible with learning rate scheduling we used in our case where the total number of steps is determined before training. 

This causes two issues in our evaluation in Figure \ref{fig:vgg-resnet}: ($i$) Although the boosting tickets reach a relatively high validation accuracy much earlier than the winning ticket, the training procedure is then hindered by the large learning rate. After the learning rate drops, the performance gap between boosting tickets and winning tickets becomes negligible. As a result, the learning rate scheduling obscures the improvement on convergence rates of boosting tickets; ($ii$) Due to fast convergence, boosting tickets tend to overfit, as observed in ResNet-18 after 50 epochs.

\begin{figure*}[ht]
    \centering
    \includegraphics[width=0.95\textwidth]{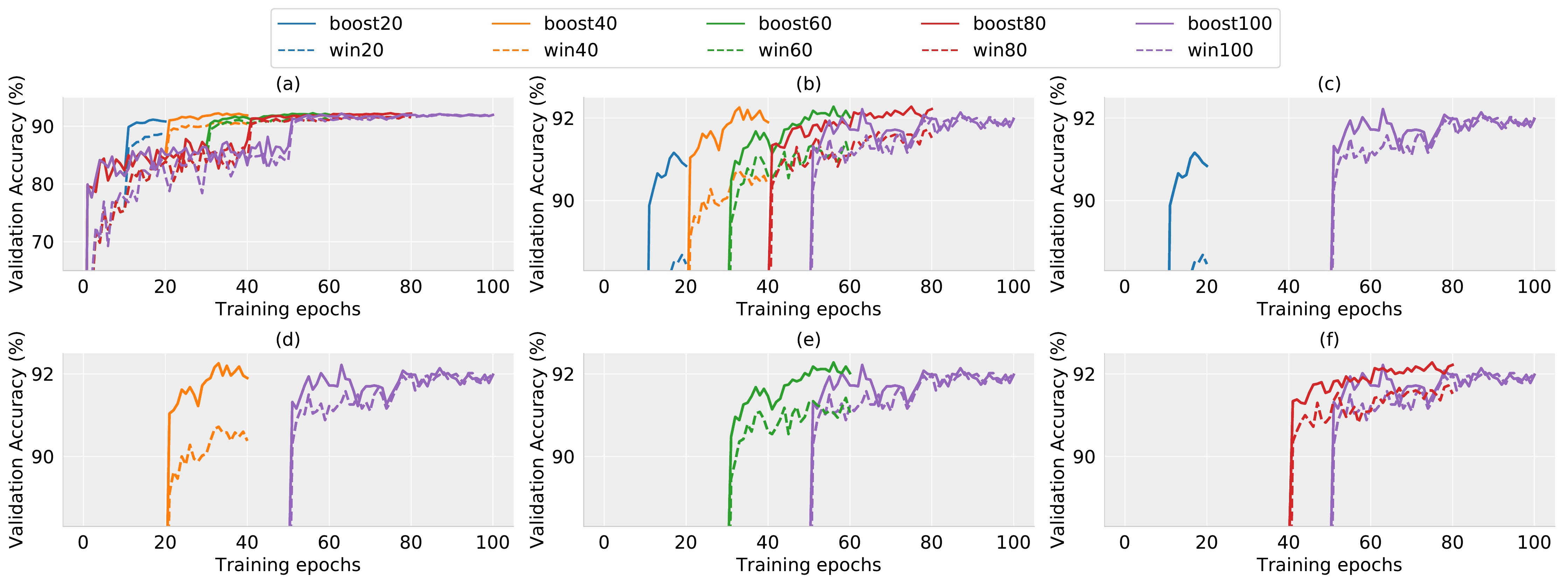}
    \caption{Validation accuracy when the total number of epochs are 20, 40, 60, 80, 100 for both the boosting tickets (straight lines) and winning tickets (dash lines) on VGG-16. Plot (a) and (b) contains the validation accuracy for all the training epochs in different scales. Plot (c,d,e,f) compare the validation accuracy between models trained for fewer epochs and the one for 100 epochs.}
    \label{fig:convergence}
\end{figure*}

To mitigate these two issues without excluding learning rate scheduling, we conduct another experiment where we mimic the early stopping procedure by gradually increasing the total number of epochs from 20 to 100. The learning rate is still dropped at the $50\%$ and $75\%$ stage. In this way, we can better understand the speed of convergence without worrying about overfitting even with learning rate scheduling involved. In figure \ref{fig:convergence}, we compare the boosting tickets and winning tickets in this manner on VGG-16.

While the first two plots in Figure \ref{fig:convergence} show the general trend of convergence, the improvement of convergence rates is much clearer in the last four plots. In particular, the validation accuracy of boosting tickets after 40 epochs is already on pair with the one trained for 100 epochs. Meanwhile, the winning tickets fall much behind the boosting tickets until 100 epochs where two finally match. 

We further investigate the test accuracy at the end of training for boosting and winning tickets in Table \ref{T:test_accuracy}. We find the test accuracy of winning tickets gradually increase as we allow for more training steps, while the boosting tickets achieve the highest test accuracy after 60 epochs and start to overfit at 100 epochs.

\begin{table}[ht]
    \caption{Final test accuracy of winning tickets and boosting tickets trained in various numbers of epochs on VGG-16.}
    \label{T:test_accuracy}
    \begin{center}
        \begin{tabular}{ c || ccccc}
        \hline
        \# of Epochs& 20 & 40 & 60 & 80 & 100\\\hline
        Winning (\%)& 88.10 & 90.03 & 90.96 & 91.79&92.00\\
        Boosting (\%)& 91.25 & 91.84 &92.13 & 92.14&92.05 \\\hline
        \end{tabular}
    \end{center}
\end{table}

Summarizing the observations above, we confirm the existence of boosting tickets and state the boosting ticket hypothesis: 

\textit{A randomly initialized dense neural network contains a sub-network that is initialized such that, when trained in isolation, converges faster than the original network and other winning tickets while matches their performance.}

In the following sections, we explain the intuition of boosting tickets and investigate three major components that affect the boosting effects.

\subsection{Intuition}

 If we think of the training procedure as a searching process from the initial weights to the optimal point in the parameter space (full path), the training procedure of the pruned model is essentially a path in a projected subspace with pruned parameters being 0 (sub-path). Suppose we want the sub-path to reach the optimal point, it is essential to follow the projection of the full path on the subspace. We follow this intuition, realizing that a smaller learning rate at the initial stage would help the sub-path follow the projected full path by avoiding large deviation. For the same reason, using the same learning rate at the initial stages for both paths also helps align them. In this way, the sub-path can quickly find the correct direction after the initial stage and start to discover a shortcut to the optimal point, resulting in boosting effects. 

\begin{figure}[t]
    \centering
    \includegraphics[width=0.5\textwidth]{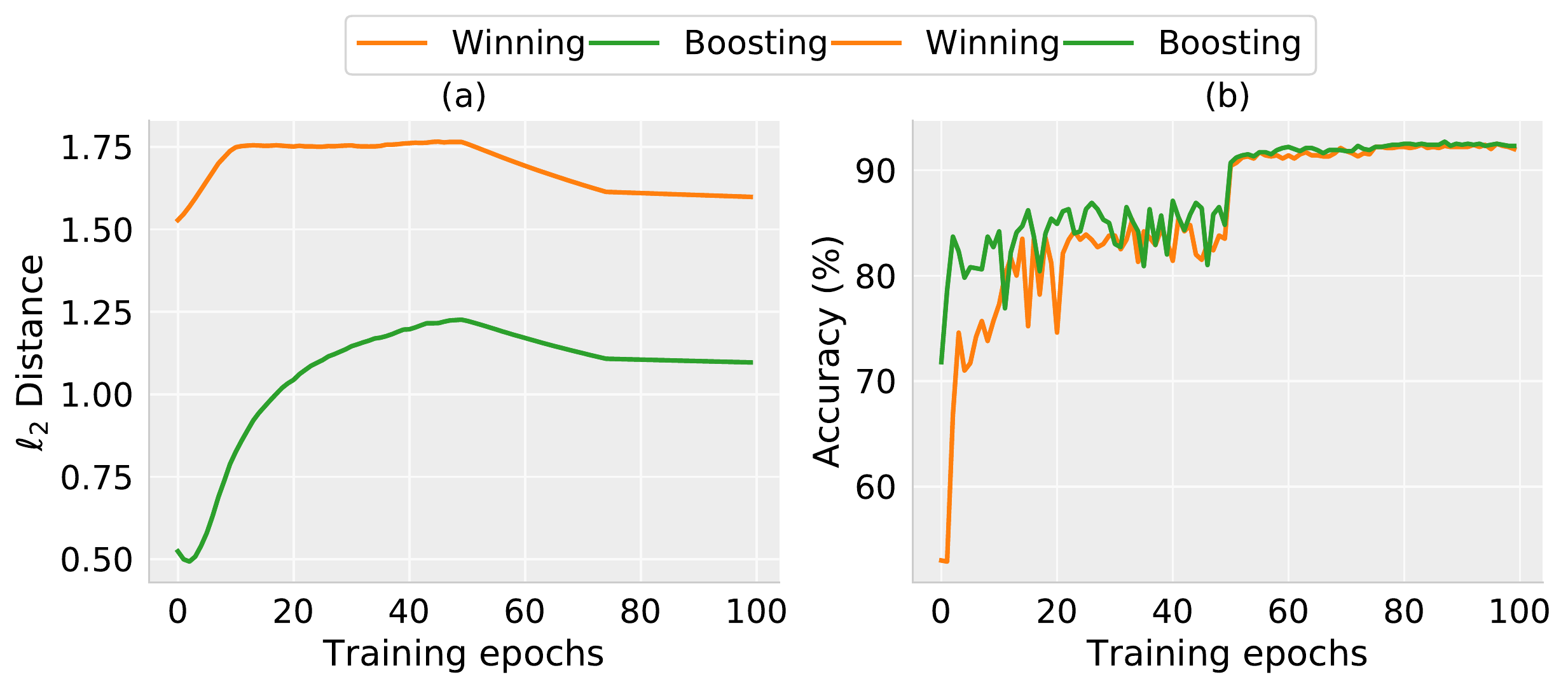}
    \caption{On VGG-16, the relative $\ell_2$ distance (a) and the corresponding accuracy (b) between the full models and pruned models .}
    \label{fig:intuition}
\end{figure}

In Figure \ref{fig:intuition}a, we calculate the relative $\ell_2$ distance between the full and the pruned model weights $w_f$ and $w_p$ generated using winning and boosting tickets $\ell_2=\frac{\|w_f-w_p\|_2}{\|w_f\|_2}$. The corresponding accuracy is reported in Figure~\ref{fig:intuition}b. It is apparent that the distance of boosting tickets is much smaller. A recent paper~\cite{evci2019difficulty} also confirmed that a linear path, although hard to find, exists between the initialization and the optimal point for pruned neural networks, which supported our intuitions about boosting effects.

\subsection{Learning Rate}
\begin{figure}
    \centering
    \includegraphics[width=0.48\textwidth]{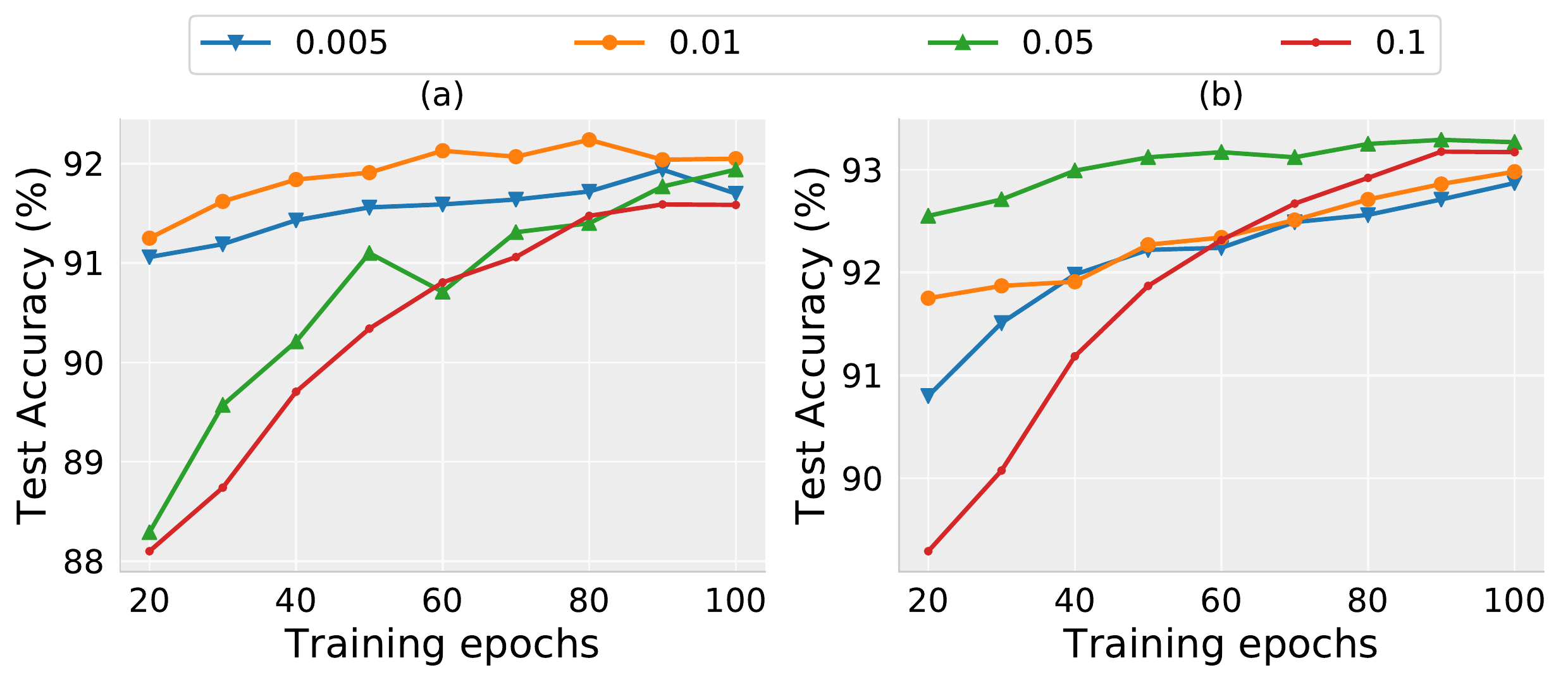}
    \caption{The final test accuracy achieved when total number of epochs vary from 20 to 100 on four different tickets. Each line denotes one winning ticket found by learning rate 0.005, 0.01, 0.05, and 0.1 for VGG-16 (a) and ResNet-18 (b).}
    \label{fig:learning_rate}
\end{figure}

As finding boosting tickets requires alternating learning rates, it is natural to assume the performance of boosting tickets relies on the choice of learning rate. Thus, we extensively investigate the influence of various learning rates.

 We use similar experimental settings in the previous section, where we increase the total number of epochs gradually and use the test accuracy as a measure of convergence rates. We choose four different learning rates 0.005, 0.01, 0.05 and 0.1 for pruning to get the tickets. All of the tickets found by those learning rates obtain the accuracy improvement over randomly reinitialized sub-model and thus satisfy the definition of winning tickets (i.e., they are all winning tickets).

As shown in the first two plots of Figure~\ref{fig:learning_rate}, tickets found by smaller learning rates tend to have stronger boosting effects. For both VGG-16 and ResNet-18, the models trained with learning rate 0.1 show the least boosting effects, measured by the test accuracy after 20 epochs of training. On the other hand, training with too small learning rate will compromise the eventual test accuracy at a certain extent. Therefore, we treat the tickets found by learning rate 0.01 as our boosting tickets for VGG-16, and the one found by learning rate 0.05 as for ResNet-18, which converge much faster than all of the rest while achieving the highest final test accuracy.

\subsection{Pruning Ratio}
\begin{figure}[ht]
    \centering
    \includegraphics[width=0.48\textwidth]{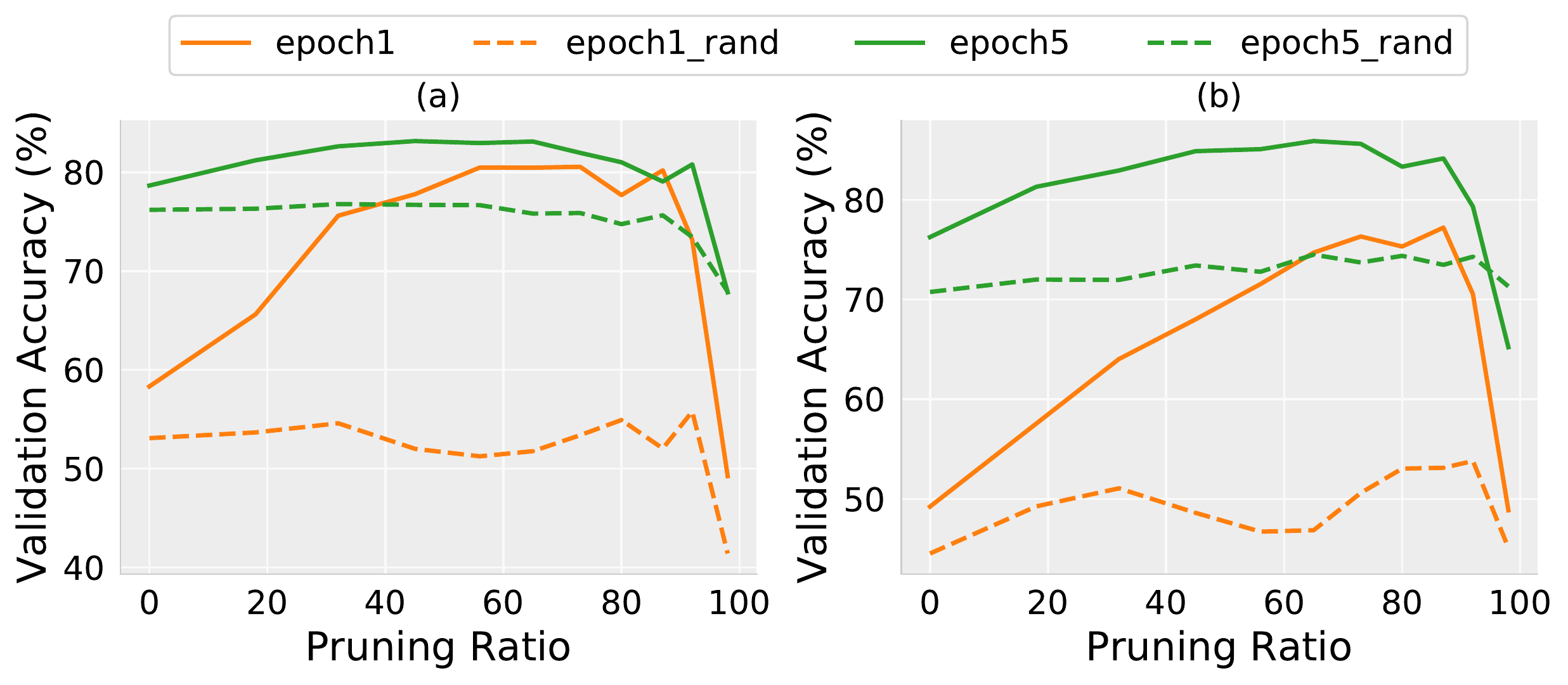}
    \caption{Under various pruning ratios, the changes of validation accuracy after the first and fifth training epoch, trained from the original initialized weights of boosting tickets and randomly reinitialized ones for VGG-16 (a) and ResNet-18 (b).}
    \label{fig:pruning_ratio}
  \end{figure}

Pruning ratio has been an important component for winning tickets \cite{frankle2018the}, and thus we investigate its effect on boosting tickets. Since we are only interested in the boosting effect, we use the validation accuracy at early stages as a measure of the strength of boosting to avoid drawing too many lines in the plots. In Figure \ref{fig:pruning_ratio}, we show the validation accuracy after the first and fifth epochs of models for different pruning ratios for VGG-16 and ResNet-18.

For both VGG-16 and ResNet-18, boosting tickets always reach much higher accuracy than randomly reinitialized sub-models, demonstrating their boosting effects. When the pruning ratio falls into the range from 60\% to 90\%, boosting tickets can provide the strongest boosting effects which obtain around 80\% and 83\% validation accuracy after the first and the fifth training epochs for VGG-16 and obtain 76\% and 85\% validation accuracy for ResNet-18. On the other hand, the increase of validation accuracy between the first training epoch and the fifth training epoch become smaller when boosting effects appear. It indicates their convergence starts to saturate due to the large learning rate at the initial stage and is ready for dropping the learning rate.
\begin{figure}[ht]
    \centering
    \includegraphics[width=0.48\textwidth]{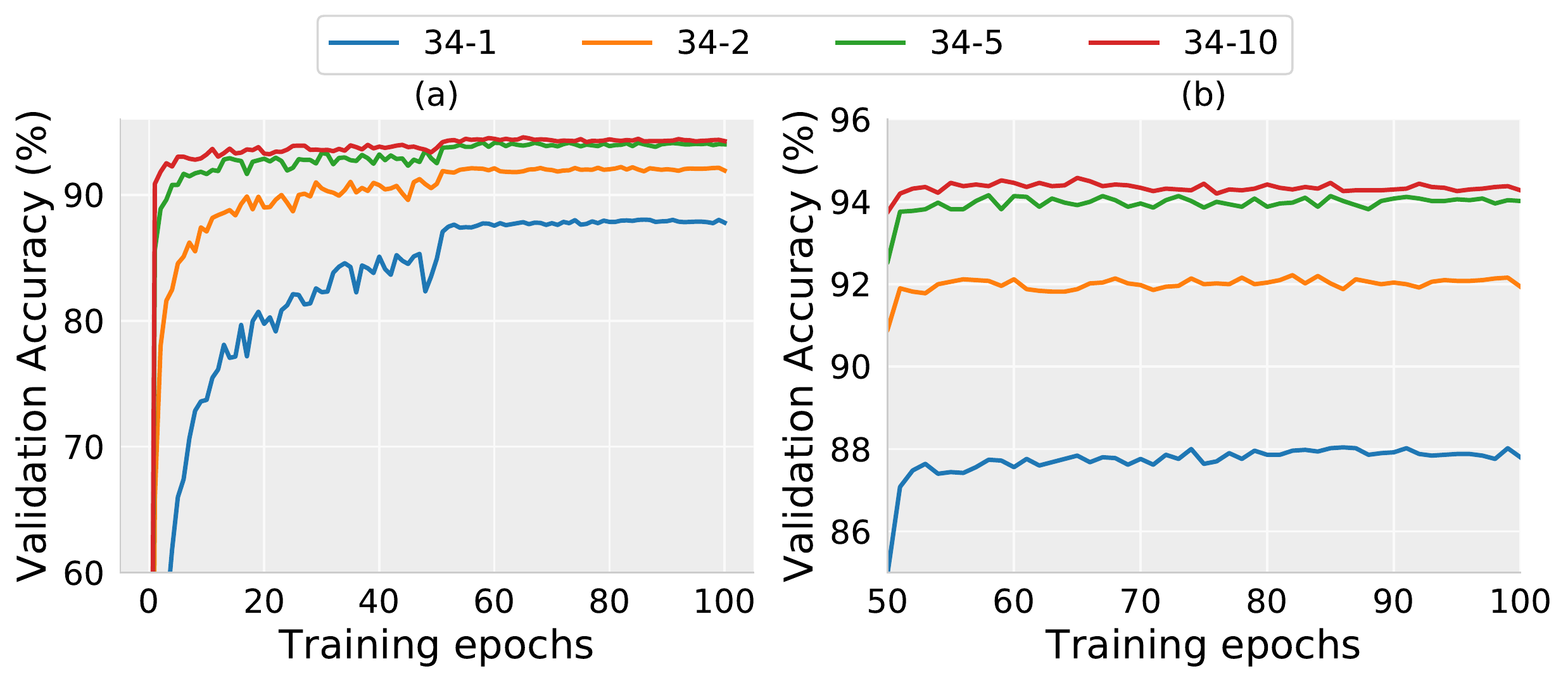}\\
    \includegraphics[width=0.48\textwidth]{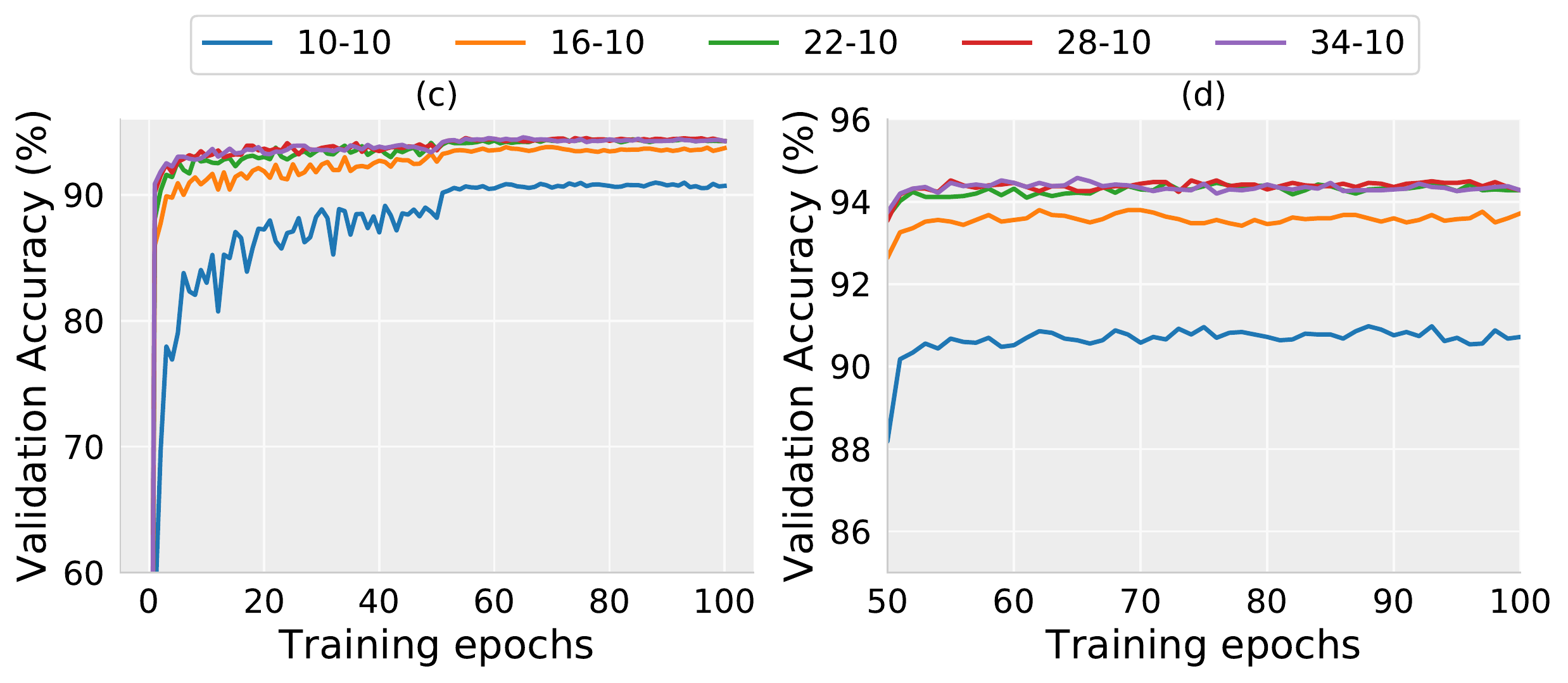}
    \caption{Plot (a) and (b) correspond to boosting tickets for various of model widths. Plot (c) and (d) correspond to boosting tickets for various of model depths. While a wider model always boosts faster, deep models have similar boosting effect when the depth is large enough.}
    \label{fig:capacity}
\end{figure}

\subsection{Model Capacity}
We finally investigate how model capacity, including the depth and width of models, affects the performance of winning tickets in the standard training setting. We use WideResNet~\cite{zagoruyko2016wide} either with its depth or width fixed and vary the other factor. In particular, we keep the depth as 34 and increases the width from 1 to 10, comparing their boosting effect. Then we keep the width as 10 and increase the depth from 10 to 34. The changes of validation accuracy of the models are shown in Figure \ref{fig:capacity}.

Overall, Figure \ref{fig:capacity} shows models with larger capacity have much better performance, though the performance keeps the same when the depth is larger than 22. Notably, we find the largest model WideResNet-34-10 achieves 90.88\% validation accuracy after only one training epoch.

\section{Lottery Ticket Hypothesis for Adversarial Training}
\label{sec:robustness}

Although the lottery ticket hypothesis is extensively studied in \cite{frankle2018the} and \cite{frankle2019lottery}, the same phenomenon in adversarial training setting lacks thorough understanding.

In this section, we show two important facts that make boosting tickets suitable for the adversarial scheme: (1) the lottery ticket hypothesis and boosting ticket hypothesis are applicable to the adversarial training scheme; (2) pruning on a weakly robust model allows to find the boosting ticket for a strongly robust model and save training cost.

Particularly, Ye et al.~\cite{ye2019second} first attempt to apply lottery ticket hypothesis to adversarial settings. However, they concluded that the lottery ticket hypothesis fails to hold in adversarial training via experiments on MNIST. In Section~\ref{subsec:ye}, our experiments demonstrate that the results they observed are not sufficient to draw this conclusion while we observe that the lottery ticket hypothesis still holds for adversarial training under more restrictive limitations.

\subsection{Applicability for Adversarial Training}
\label{subsec:adv1}
\begin{figure}[t]
    \centering
    \includegraphics[width=0.48\textwidth]{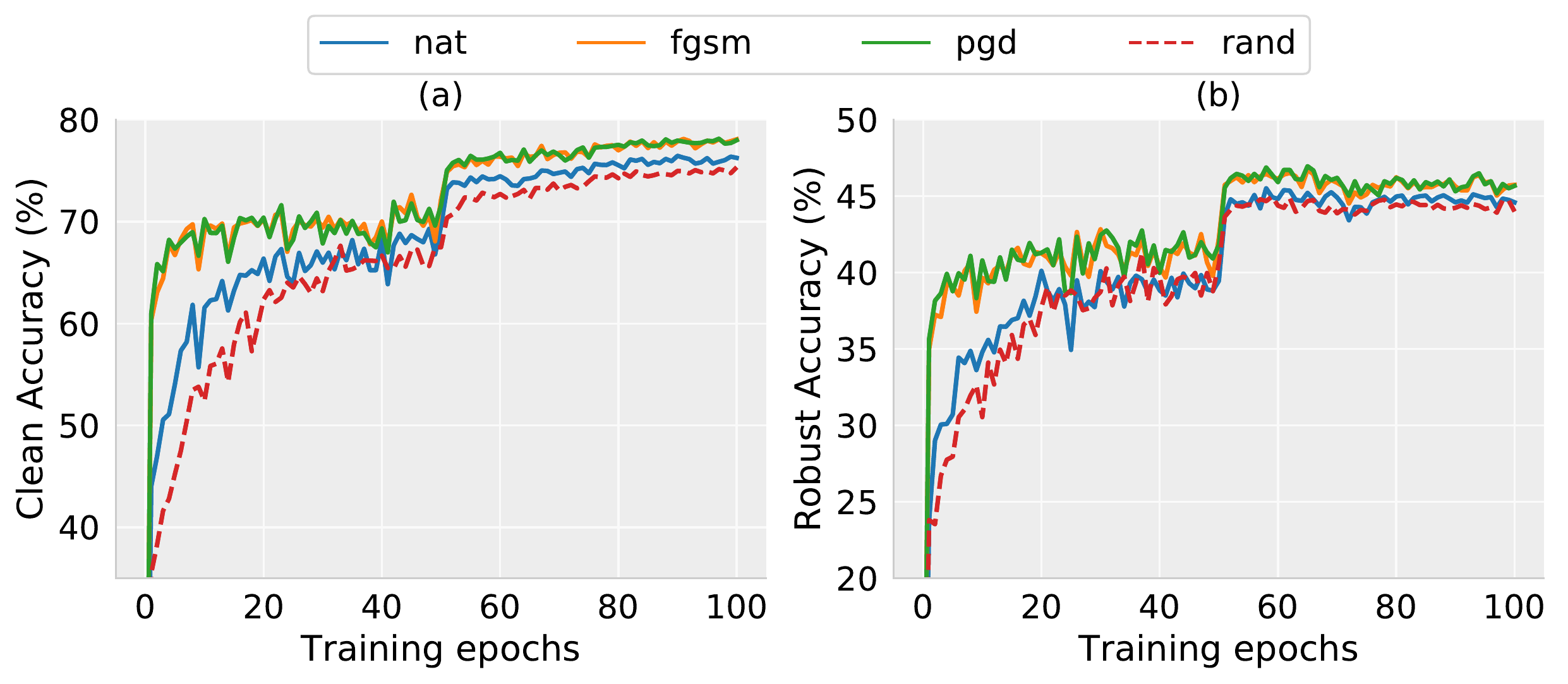}
    \caption{The clean accuracy (a) and robust accuracy (b) of pruned models on the validation set. The models are pruned based on different training methods (natural training, FGSM-based adversarial training, and PGD-based adversarial training). For each obtained boosting ticket, it is retrained with PGD-based adversarial training with 100 training epochs.}
    \label{fig:nat_adv}
\end{figure}

In the following experiment, we use a naturally trained model, that is trained in the standard manner, and two adversarially trained models using FGSM and PGD respectively to obtain the tickets by pruning these models. Then we retrain these pruned models with the same PGD-based adversarial training from the same initialization. In Figure \ref{fig:nat_adv}, we report the corresponding accuracy on the original validation sets and on the adversarially perturbed validation examples, noted as clean accuracy and robust accuracy. We further train the pruned model from random reinitialization to validate lottery ticket hypothesis.

Unless otherwise stated, in all the PGD-based adversarial training, we keep the same setting as~\cite{madry2017towards}. The PGD attacks are performed in 10 steps with step size $2/255$ (PGD-10). The PGD attacks are bounded by $8/255$ in its $\ell_\infty$ norm. For the FGSM-based adversarial training, the FGSM attacks are bounded by $8/255$.

Both models trained from the boosting tickets obtained with FGSM- and PGD-based adversarial training demonstrate superior performance and faster convergence than the model trained from random reinitialization. This confirms the lottery ticket hypothesis and boosting ticket hypothesis are applicable to adversarial training scheme on both clean accuracy and robust accuracy. More interestingly, the performance of the models pruned with FGSM- and PGD-based adversarial training are almost the same. This observation suggests it is sufficient to train a weakly robust model with FGSM-based adversarial training for obtaining the boosting tickets and retrain it with stronger attacks such as PGD.

This finding is interesting because FGSM-based adversarial trained models will suffer from label leaking problems as learning weak robustness~\cite{kurakin2016adversarial}. In fact, the FGSM-based adversarially trained model from which we obtain our boosting tickets has 89\% robust accuracy against FGSM but with only 0.4\% robust accuracy against PGD performed in 20 steps (PGD-20). However, Figure~\ref{fig:nat_adv} shows the following PGD-based adversarial retraining on the boosting tickets obtained from that FGSM-based trained model is indeed robust. Further discussions can be found in Section~\ref{sec:discussion}.
 \begin{figure*}[ht]
	\centering
	\includegraphics[width=0.96\textwidth]{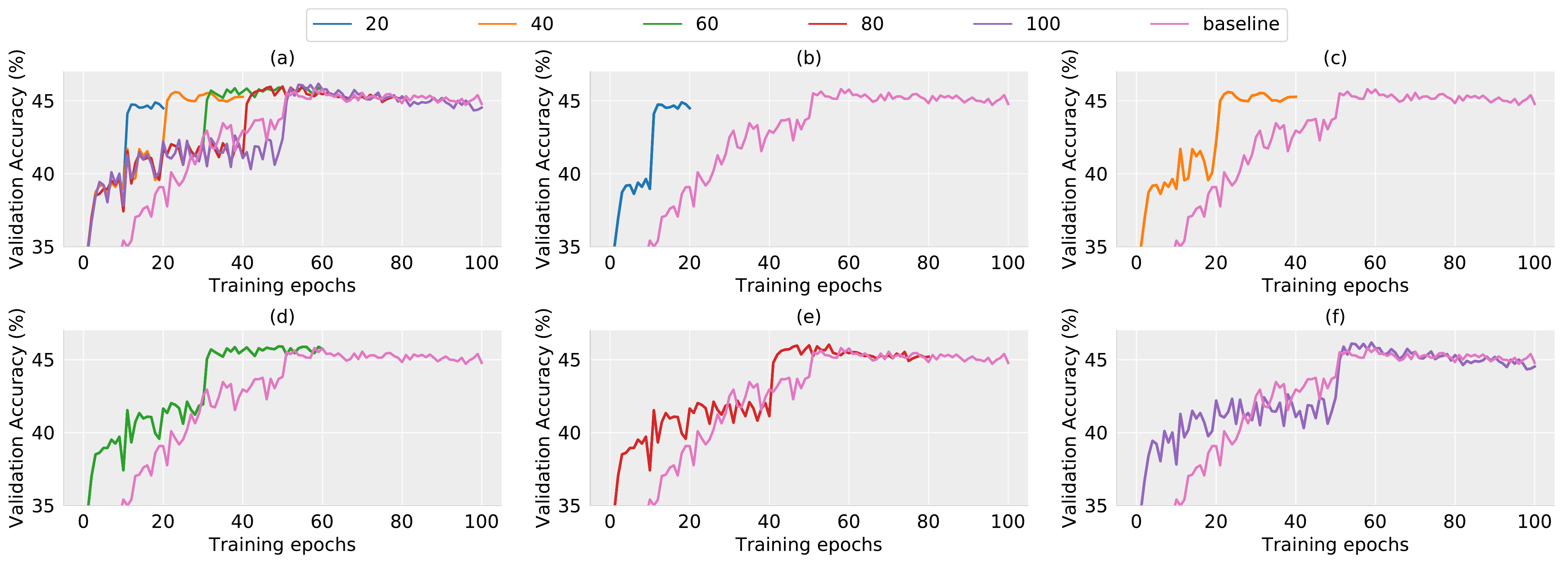}
	\caption{Validation robust accuracy of pruned models with PGD-based adversarial training on VGG-16 where the total number of epochs are 20, 40, 60, 80, 100 respectively. Plot (a) and (b) show all the results while plot (c), (d), (e), (f) compare each model with the baseline model. The baseline model is obtained by 100-epoch PGD-based adversarial training on the original full model.}
	\label{fig:adv_convergence}
	\vspace{-10pt}
\end{figure*}

\subsection{Explain the Failure in \cite{ye2019second}}
 \label{subsec:ye}
In~\cite{ye2019second}, the authors argued that the lottery ticket hypothesis fails to hold in adversarial training via experiments on MNIST. We show they fail to observe winning tickets because the models they used have limited capacity. 

\begin{figure}[ht]
	\centering
	\includegraphics[width=0.45\textwidth]{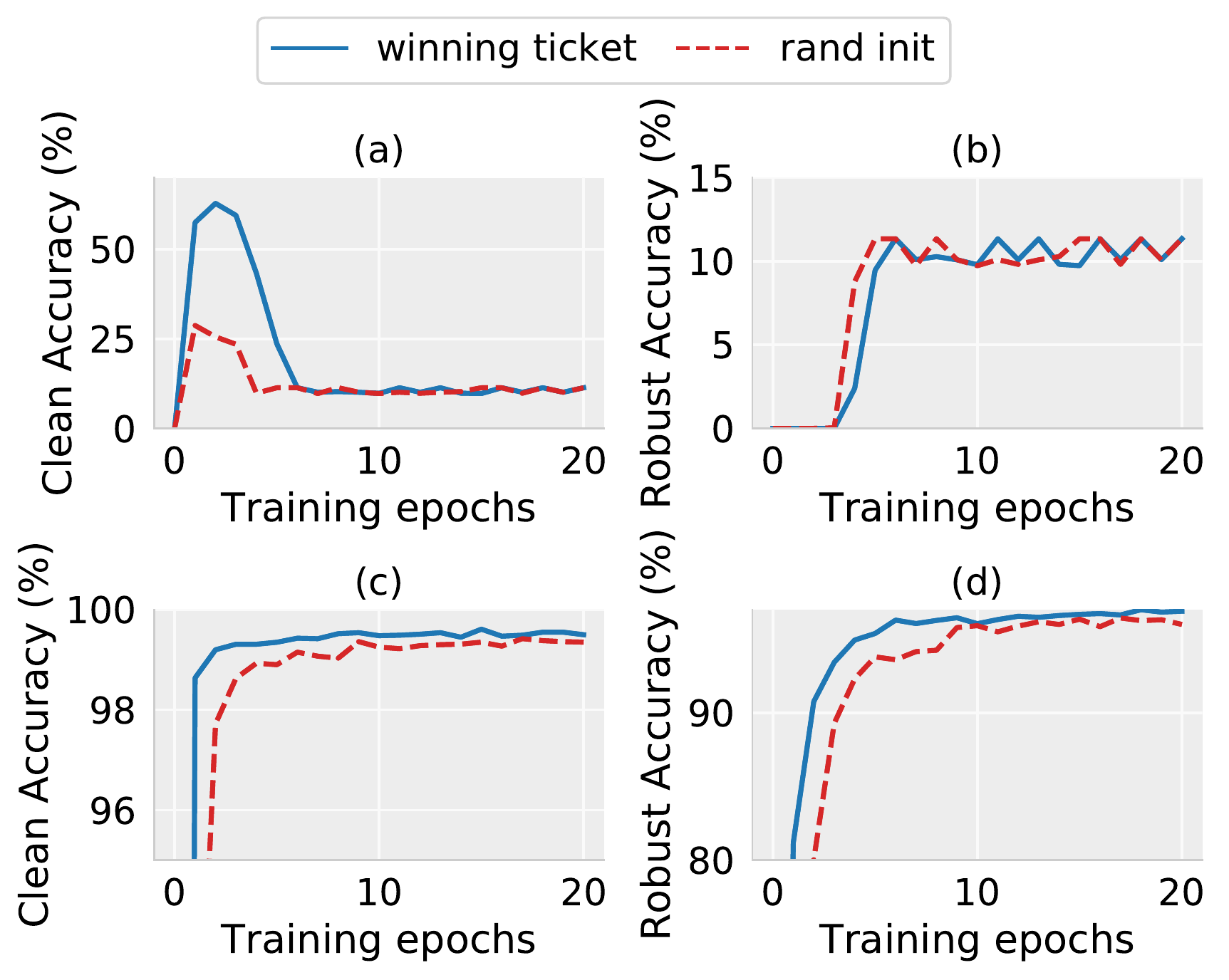}
	\caption{We show clean (a,c) and robust accuracy (b,d) for both winning tickets and randomly initialized weights on LeNet (a,b) and Vgg-16 (c,d) on MNIST with adversarial training.}
	\label{fig:adv_lenet}
\end{figure}

We first reproduce their results to show that, in the adversarial setting, small models such as a CNN with two convolutional layers used in \cite{ye2019second} can not yield winning tickets when pruning ratio is large. In Figure \ref{fig:adv_lenet}, plot (a) and (b) are the clean and robust accuracy of the pruned models when the pruning ratio is $80\%$. The pruned model eventually degrades into a trivial classifier where all example are classified into the same class with 11.42\%/11.42\% valid/test accuracy. On the other hand, when we use VGG-16, as shown in plot (c) and (d), the winning tickets are found again. This can be explained as adversarial training requires much larger model capacity than standard training, which is extensively discussed in \cite{madry2017towards}. As the result, the pruned small models become unstable during training and yields degrading performance.

As a comparison, the experimental results from Figure \ref{fig:nat_adv} indicate when larger models are used, the lottery ticket hypothesis still applies to the adversarial trainig scheme.

\subsection{Convergence Speedup}

We then investigate how boosting tickets can accelerate the adversarial training procedure by conducting the same experiments as in Figure \ref{fig:convergence} but in the adversarial training setting. The results for validation accuracy and test accuracy are presented in Figure \ref{fig:adv_convergence} and Table \ref{T:fgsm_test_accuracy} respectively. 

In Figure~\ref{fig:adv_convergence}, all the training plots are PGD-based adversarial training on the same boosting ticket and learning rate scheduling but with different training epochs. We follow the same procedure described in Section~\ref{subsec:adv1} to obtain the boosting ticket. Specifically, we train VGG-16 model with 100-epoch FGSM-based adversarial training and then prune 80\% of the weight connections. From Figure~\ref{fig:adv_convergence}, we can see it is sufficient to train 60 epochs to achieve similar robust accuracy as the full model trained for 100 epochs on our boosting ticket.

Also, in Table~\ref{T:fgsm_test_accuracy}, we have compared the original full models trained by 100-epoch PGD-based adversarial training with the ones trained on our boosting ticket with different epochs. In general, our models trained on boosting ticket can obtain at most $0.5\%$ higher robust accuracy and regular accuracy than the original one. It indicates (1) the lottery ticket holds for adversarial training as well and (2) our boosting ticket can still enjoy the benefits of both lottery ticket hypothesis and convergence speedup.

\begin{table}[ht]
	\caption{Best test clean and robust accuracy for PGD-based adversarial training on boosting tickets obtained by FGSM-based adversarial training in various numbers of epochs on VGG-16. Baseline model is obtained by 100-epoch PGD-based adversarial training on original full model.}
	\vspace{-10pt}
	\tabcolsep=3.5pt
	\label{T:fgsm_test_accuracy}
	\begin{center}
		\begin{tabular}{ c || ccccccc}\hline
			\# of Epochs& 20 & 40 & 60 & 80  & 100 & Baseline\\\hline
			Robust Acc.& 44.49 & 45.27 & \textbf{45.73} & 45.20 & 44.53 & 44.78\\
			Clean Acc.& 75.15 & 76.28 & 76.48 & 77.60 & \textbf{78.07} & 77.21\\\hline
		\end{tabular}
	\end{center}
	\vspace{-10pt}
\end{table}

\subsection{Boosting Ticket Applications on adversarially trained WideResNet-34-10}
\label{subsec:wrn}
Until now, we have confirmed that boosting tickets exist consistently across different models and training schemes and convey important insights on the behavior of pruned models. However, in the natural training setting, although boosting tickets provide faster convergence, it is not suitable for accelerating the standard training procedure as pruning to find the boosting tickets requires training full models beforehand. On the other hand, the two observations mentioned in Section~\ref{sec:robustness} enable boosting tickets to accelerate adversarial training. 

In Table \ref{T:robustness}, we apply adversarial training to WideResNet-34-10, which has the same structure used in \cite{madry2017towards}, with the proposed approach for 40, 70 and 100 epochs and report the best accuracy/robust accuracy under various attacks among the whole training process. In particular, we perform 20-step PGD, 100-step PGD as white-box attacks where the attackers have the access to the model parameters.

It might be suspicious if the resulting models from pruning and adversarial training are indeed robust against strong attacks, as the pruning mask is obtained from a weakly robust model. We conduct extensive experiments on CIFAR-10 with WideResNet-34-10 to evaluate the robustness of this model and compare to the robust model trained with Madry et al's method~\cite{madry2017towards}. Therefore, we include results for C\&W attacks \cite{carlini2017towards} and transfer attacks \cite{papernot2016transferability,liu2016delving} where we attack one model with adversarial examples found by 20-step PGD based on other models. 

We find the adversarial examples generated from one model can transfer to another model with a slight decrease on the robust error. It indicates our models and Madry et al's models share adversarial examples and further share decision boundaries.
\begin{table}[ht]
    \caption{Best test clean accuracy (the first row), robust accuracy (the second to fourth rows), transfer attack accuracy (the middle four rows), and training time for PGD-based adversarial training (the last four rows) on boosting tickets obtained by FGSM-based adversarial training in various of numbers of epochs on WideResNet-34-10. Overall, our adversarial training strategy based on boosting tickets is able to save up to 49\% of the total training time while achieving higher robust accuracy compared to the regular adversarial training on the original full model.}
    \tabcolsep=3.5pt
    \vspace{-10pt}
    \label{T:robustness}
    \begin{center}
        \begin{tabular}{ c || rrrr}
        \hline
        &\multicolumn{4}{c}{Test Accuracy(\%)}\\ 
        \cline{1-5}
        Models & Madry's & Ours-40 & Ours-70 & Ours-100\\\hline
          Natural& 86.21 & 87.72 & \textbf{87.85} & 87.35 \\
          PGD-20 & 50.07 & 50.37 & \textbf{50.48} & 49.92 \\
          PGD-100 & 49.32 & 49.28 &\textbf{49.58} & 49.11\\
          C\&W & 50.46 & \textbf{50.92} & 50.82 & 50.37 \\\hline
          Madry's &-&58.16&57.39&57.63\\
          Ours-40 &58.69&-&54.04&56.11\\
          Ours-70 &58.77&54.60&-&55.23\\
          Ours-100 &58.61&56.62&55.20&-\\\hline
          Pruning Time(s) & 0 &15,462&15,462&15,462\\
          Training Time(s)& 134,764&54,090&94,796&137,105\\
          Total Time(s) &134,764&\textbf{69,552}&110,258&152,567\\
          Ours/Madry's & - & \textbf{0.51} & 0.82 & 1.13 \\          \hline
        \end{tabular}
    \end{center}
    \vspace{-20pt}
\end{table}

We report the time consumption for training each model to measure how much time is saved by boosting tickets. We run all the experiments on a workstation with 2 V100 GPUs in parallel. From Table \ref{T:robustness} we observe that while our approach requires pruning before training, it is overall faster as it uses FGSM-based adversarial training. In particular, to achieve its best robust accuracy, original Madry et al.'s training method~\cite{madry2017towards} requires 134,764 seconds on WideResNet-34-10. To achieve that, our boosting ticket only requires 69,552 seconds, including 15,462 seconds to find the boosting ticket and 54,090 seconds to retrain the ticket, saving 49\% of the total training time.

\section{Discussion and Future Work}
\label{sec:discussion}

\noindent\textbf{Not knowledge distillation.} It may seem that winning tickets and boosting tickets behave like knowledge distillation \cite{ba2014deep,hinton2015distilling} where the learned knowledge from a large model is transferred to a small model. This conjecture may explain the boosting effects as the pruned model quickly recover the knowledge from the full model. However, the lottery ticket framework seems to be distinctive to knowledge distillation. If boosting tickets simply transfer knowledge from the full model to the pruned model, then an FGSM-based adversarially trained model should not find tickets that improves the robustness of the sub-model against PGD attacks, as the full model itself is vulnerable to PGD attacks. Yet in Section~\ref{subsec:adv1} we observe an FGSM-based adversarially trained model still leads to boosting tickets that accelerates PGD-based adversarial training. We believe the cause of boosting tickets requires further investigation in the future.

\noindent\textbf{Accelerate adversarial training.} Recently, \cite{shafahi2019adversarial} propose to reduce the training time for PGD-based adversarial training by recycling the gradients computed for parameter updates and constructing adversarial examples. While their approach focuses on reducing the computational time for each epoch, our method focuses more on the convergence rate (i.e., reduce the number of epochs required for convergence). Therefore, our approach is compatible with theirs, making it a promising future direction to combine both to further reduce the training time.

\section{Conclusion}

\label{sec:conclusion}
In this paper, we investigate boosting tickets, sub-networks coupled with certain initialization that can be trained with significantly faster convergence rate. As a practical application, in the adversarial training scheme, we show pruning a weakly robust model allows to find boosting tickets that can save up to 49\% of the total training time to obtain a strongly robust model that matches the state-of-the-art robustness. 
Finally, it is an interesting direction to investigate whether there is a way to find boosting tickets without training the full model beforehand, as it is technically not necessary.


{\small
\bibliographystyle{ieee_fullname}
\bibliography{egbib}
}


\appendix


\section{Model Architectures and Setup} 
\label{appd:model}

In Table~\ref{T:models}, we summarize the number of parameters and parameter sizes of all the model architectures that we evaluate with including VGG-16~\cite{simonyan2014very}, ResNet-18~\cite{he2016deep}, and the variance of WideResNets~\cite{zagoruyko2016wide}.

\begin{table}[ht]
    \caption{Number of parameters and parameter sizes for various architectures.}
    \label{T:models}
    \begin{center}
        \begin{tabular}{ l || r|r}
        \hline
          & \# of Parameters & Size (MB)\\\hline
         VGG-16 & 29,975,444 & 114.35\\\hline
         ResNet-18 & 11,173,962 & 42.63 \\\hline
         WideResNet-34-10 & 46,160,474 & 176.09 \\
         WideResNet-28-10 & 36,479,194 & 139.16 \\
         WideResNet-22-10 & 26,797,914 & 102.23 \\
         WideResNet-16-10 & 17,116,634 & 65.29 \\
         WideResNet-10-10 & 7,435,354 & 28.36 \\
         WideResNet-34-5 & 11,554,074 & 44.08 \\
         WideResNet-34-2 & 1,855,578 & 7.08 \\
         WideResNet-34-1 & 466,714 & 1.78 \\\hline
         
        \end{tabular}
    \end{center}
\end{table}

\section{One Shot Pruning}
\label{appd:ip}

\begin{figure}[ht]
	\centering
	\includegraphics[width=0.45\textwidth]{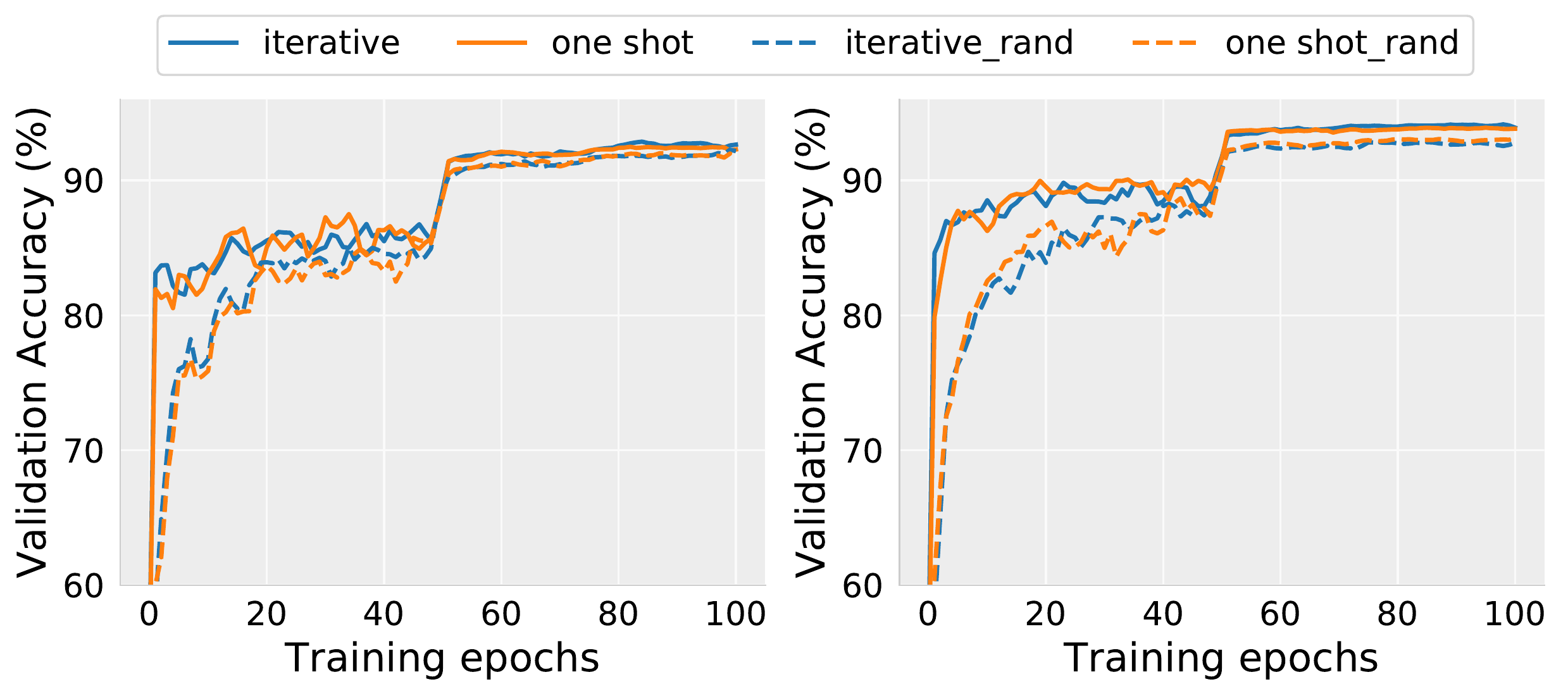}
	\caption{We compare tickets obtained via iterative pruning and one shot pruning on VGG-16 (left) and ResNet-18 (right). We plot the validation accuracy of models from both approaches and the corresponding randomly initialized models.}
	\label{fig:iterative}
\end{figure}

In Figure \ref{fig:iterative}, we track the training of models obtained from both iterative pruning and one shot pruning. We find the performance of both, in terms of the boosting effects and final accuracy, is indistinguishable.

\section{Experiments on MNIST}

In this section, we report experiment results on MNIST for the standard setting, where we use LeNet with two convolutions and two fully connected layers for the classification task.

As for MNIST we do not use learning rate scheduling, early stopping is then used to determine the speed of convergence. In Table \ref{T:mnist_nat}, we report the epochs when early stopping happens and the test accuracy to illustrate the existence of boosting tickets for MNIST. While winning tickets converge at the 18th epoch, boosting tickets converge at the 11th epoch, indicating faster convergence.
\begin{table}[ht]
	\caption{The epochs when early stopping happens and the corresponding accuracy for the full model, winning tickets, boosting tickets, and randomly initialized model based on LeNet with two convolutional layers and two fully connected layers.}
	\tabcolsep=3.5pt
	\label{T:mnist_nat}
	\begin{center}
		\begin{tabular}{ c || cccc}
			\hline
			&Full Model & Winning&Boosting&Rand Init\\\hline
			Early Stopping& 20 & 18 & 11 & 16  \\
			Test Accuracy & 99.18 & 99.24 & 99.23 & 98.97\\\hline
		\end{tabular}
	\end{center}
\end{table}



\end{document}